\documentclass{article}

\usepackage[preprint]{main}

\usepackage[T1]{fontenc}    
\usepackage{hyperref}       
\usepackage{url}            
\usepackage{booktabs}       
\usepackage{amsfonts}       
\usepackage{nicefrac}       
\usepackage{microtype}      
\usepackage{xcolor}         

\usepackage{graphicx}
\usepackage{amsmath}
\usepackage{float} 

\usepackage{subfigure}

\title{HYDEN: Hyperbolic Density Representations for Medical Images and Reports}

\author{%
Zhi Qiao$^{1}$ \quad Linbin Han$^{1,2}$ \quad Xiantong Zhen$^1$ \quad Jia-Hong Gao$^2$ \quad Zhen Qian$^1$ \\
$^1$Institute of Intelligent Diagnostics, Beijing United-Imaging Research Institute of Intelligent Imaging \\ $^2$School of Physics, Peking University\\
}

\begin{document}

\maketitle

\begin{abstract}
In light of the inherent entailment relations between images and text, hyperbolic point vector embeddings, leveraging the hierarchical modeling advantages of hyperbolic space, have been utilized for visual semantic representation learning. However, point vector embedding approaches fail to address the issue of semantic uncertainty, where an image may have multiple interpretations, and text may refer to different images, a phenomenon particularly prevalent in the medical domain. Therefor, we propose \textbf{HYDEN}, a novel hyperbolic density embedding based image-text representation learning approach tailored for specific medical domain data. This method integrates text-aware local features alongside global features from images, mapping image-text features to density features in hyperbolic space via using hyperbolic pseudo-Gaussian distributions. An encapsulation loss function is employed to model the partial order relations between image-text density distributions. Experimental results demonstrate the interpretability of our approach and its superior performance compared to the baseline methods across various zero-shot tasks and different datasets.
\end{abstract}

\section{Introduction}
In recent years, cross-modal text-image representation learning has achieved tremendous success and drawn widespread attention in many tasks such as zero-shot learning and image-text retrieval. This success is largely due to the use of large volumes of weakly-supervised image-text pair data to enhance vision-language representation learning \citep{radford2021clip}. In the field of medical imaging, cross-modal representation learning tailored to specific domain data, such as chest radiographs and their associated radiology reports, can yield robust and powerful foundation models in specialized areas \citep{medicalclipchallenges}.

As the proverb goes, 'A picture is worth a thousand words.' This suggests that an image inherently contains more information than a textual description of it, which can be seen as merely a simplified abbreviation of the image. This relationship, where the text may serve as an entailment of the image, can be considered as visual-semantic hierarchy \citep{vendrov2016orderembeddings}. Consequently, it is a plausible hypothesis that incorporating such inductive biases of visual semantic hierarchies into cross-modal alignment tasks could enhance the generalizability of representations and improve the interpretability of learning representations. \citet{vendrov2016orderembeddings} introduced an order embedding strategy considering these hierarchical semantic during the text-image alignment process. However, numerous studies \citep{PoincarEmbeddings,ContinuousLorentzModel,HyperMiner,Hyperbolicclass,Hyperbolicgraph} have demonstrated that modeling data with inherent hierarchical features in non-Euclidean hyperbolic spaces can provide superior representations. By leveraging the advantages of hyperbolic space in modeling hierarchical structures and the generalization capabilities of cross-modal contrastive learning in zero-shot scenarios, \cite{hyperclip} has proposed cross-modal hyperbolic representation learning. This approach employs the Lorentz manifold to map both image and text features into hyperbolic space, utilizing angular constraints based on entailment to learn the hierarchical order between text and images.

\begin{figure}[htbp]	
	\subfigure[ ] 
	{
		\begin{minipage}{5cm}
			\centering          
			\includegraphics[width=6cm]{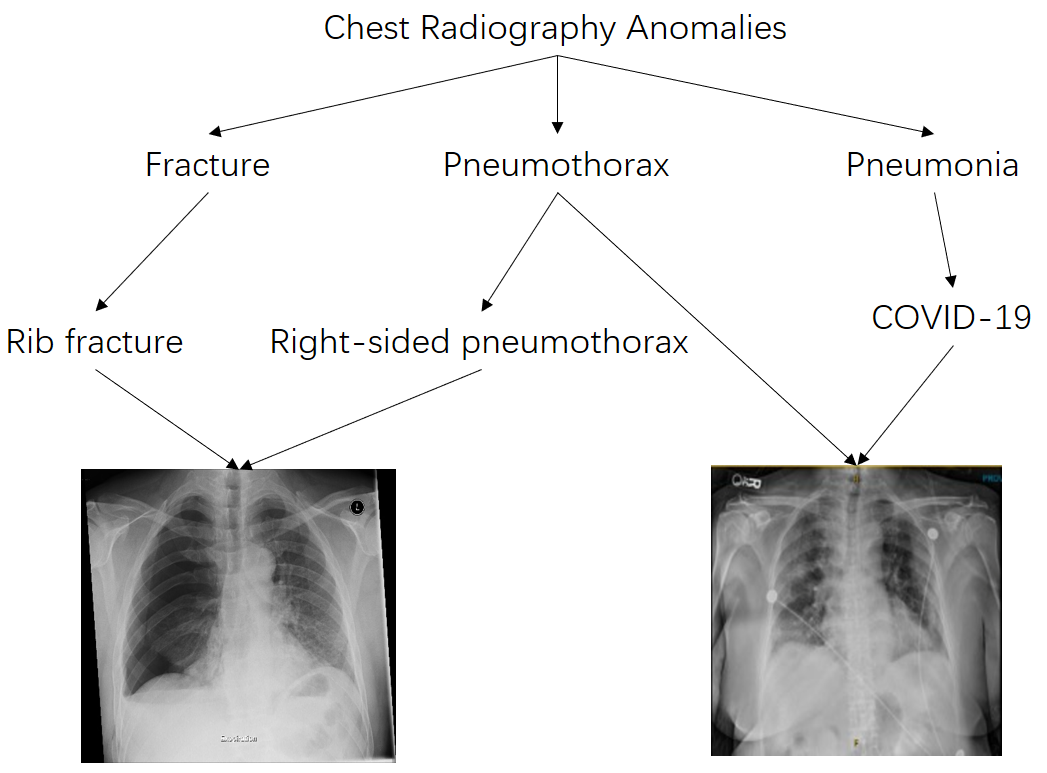}   
		\end{minipage}
	}
\hspace{1cm}
	\subfigure[ ] 
	{
		\begin{minipage}{6cm}
			\centering     
			\includegraphics[width=7cm]{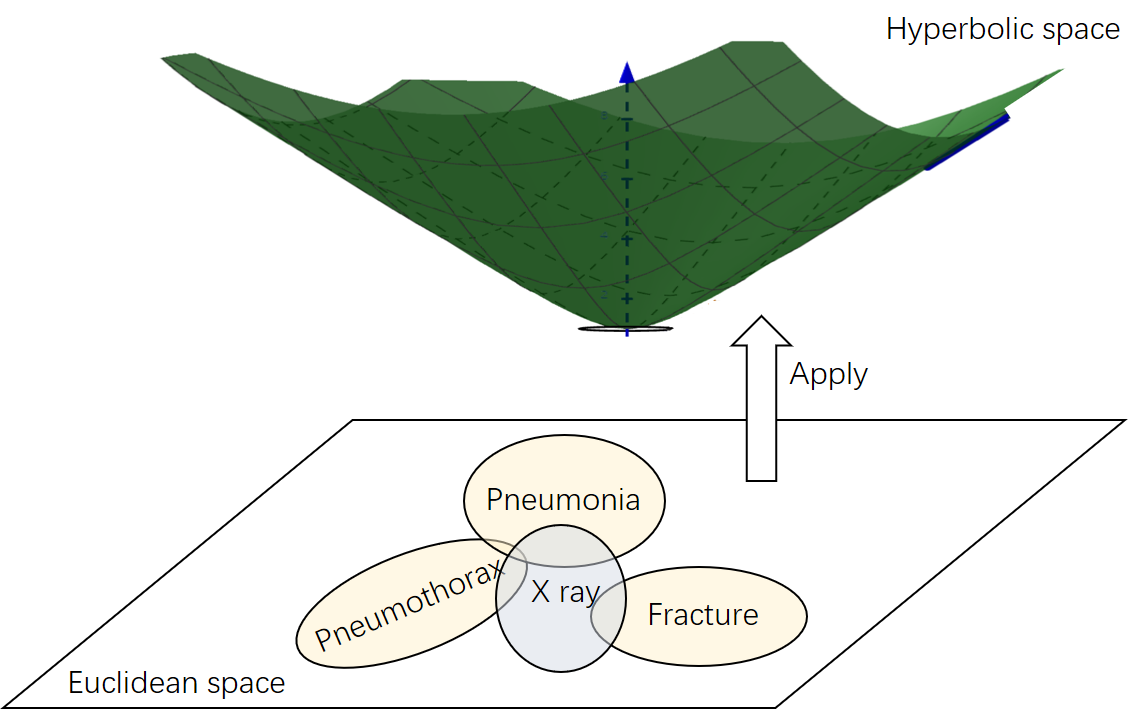}   
		\end{minipage}
	}

    \caption{
(a) Depiction of the visual-semantic hierarchy in the medical text-image domain, illustrating how different medical concepts are organized and interconnected with each other and with medical images. (b) Representation of medical data embeddings transitioning from Euclidean to hyperbolic space to effectively capture and represent the density partial ordering, while maintaining the integrity of relative density relationships.
}
    \label{fig:motivation}  
\end{figure}

However, representing image-text with point vectors has a clear limitation: it cannot express semantic uncertainty \citep{gaussianembedding,gaussianembedding_concept}, meaning that a single image can generate different descriptions from various perspectives, and similarly, a single textual description can describe different but related images. This phenomenon is particularly evident in medical imaging and radiology reports. For instance, as depicted in Fig.\ref{fig:motivation}(a), consider a patient with a rib fracture suspected of having right-sided pneumothorax. In the radiology report for this patient, the physician describes the imaging findings related to pneumothorax, highlighting the presence of a white line around the visceral pleural edge. Clinically, numerous pulmonary diseases, such as tuberculosis, cystic fibrosis, and pneumocystis jiroveci pneumonia, predispose individuals to pneumothorax. In domains such as document embedding \citep{docummentembedding} and graph embedding \citep{auhorembedding}, the utilization of probability density embedding to represent objects as distributions within the target space effectively addresses this semantic uncertainty, resulting in significantly improved performance compared to point vector embedding.



Building on the motivation outlined above, which focuses on the hierarchical visual semantic features and inherent semantic uncertainty in medical imaging, we propose \textbf{HYDEN}, a hyperbolic density representations for medical images and reports. This approach leverages the advantages of hyperbolic space for capturing visual-semantic hierarchy, while incorporating a probability density embedding strategy to model semantic uncertainty. The main contributions are as follows:



\begin{itemize}
\itemsep0em
\item To the best of our knowledge, this is the first model to apply cross-modal representation learning to medical image-text data within hyperbolic space.
\item We introduce a text-aware image local feature extraction method that focuses on local regions, enhancing the granularity of analysis; moreover, we employ encapsulation constraints to model the density order between images and text, fostering a deeper semantic connection.
\item Extensive experiments were conducted to validate the performance of our algorithm against baseline models through both quantitative and qualitative analyses. These experiments demonstrate the superior capabilities of our approach in achieving semantic alignment.
\end{itemize}


\section{Related Work}

Image-text representation learning has garnered substantial interest due to its potential to enhance visual representation. Traditional methods predominantly employ contrastive metric learning approaches, with CLIP \citep{radford2021clip} being a notable example that has demonstrated remarkable results. These methods typically operate in the Euclidean space and have been extensively applied across various general domains.

However, the medical field presents unique challenges due to the domain-specific nature and the complex prior knowledge embedded in medical image-text data. In response, several studies have explored image-text representation learning specifically tailored to medical contexts \citep{medical_clip_cluster, medical_clip_gloria_symmetry, medical_clip_prior, gloria}. Despite their advancements, these approaches continue to operate within the confines of Euclidean space. The inherent hierarchical semantics between images and texts, particularly pronounced in medical datasets, suggest that hyperbolic space could offer significant advantages. Hyperbolic space naturally accommodates hierarchical data structures, making it a compelling alternative for modeling complex semantic relationships. Building on this premise, the MERU framework introduced hyperbolic image-text embedding \citep{hyperclip}, representing a significant departure from traditional Euclidean methods.

Building upon the concept introduced by MERU, our work proposes a novel approach by integrating density embedding to capture semantic uncertainty, which is a feature not adequately addressed by point vector embeddings. While density embedding has been previously utilized for capturing the uncertainty of semantics and modeling asymmetric relationships like entailment \citep{gaussianembedding, gaussianembedding_concept, gaussiangraphembedding}, these implementations have been confined to Euclidean space. Our method extends this concept into hyperbolic space.


\section{Preliminaries}
\textbf{Hyperbolic Geometry} Hyperbolic geometry is a non-Euclidean geometry with a constant negative  curvature, and it can be visualized as the forward sheet of the two-sheeted hyperboloid. In this study, we will use the Lorentz model on the upper half of a two-sheeted hyperboloid, as claimed in \citep{ContinuousLorentzModel}, comes with a simpler closed form of the geodesics and does not suffer from the numerical instabilities in approximating the distance. Lorentz model $\mathbb{H}^n$ processing a constant curvature $-c$ can be represented as a set of points $z\in \mathbb{R}^{n+1}$. Lets $z, z'\in \mathbb{H}^n$, the Lorentzian product $\left \langle z, z'  \right \rangle_{\mathcal{L}} = -z_0z_0^{'} + \sum_{i=1}^{n}z_iz_i^{'}$. And, $\mathbb{H}^n = \{z\in \mathbb{R}^{n+1}: \left \langle z, z \right \rangle_{\mathcal{L}}=-1/c,~c>0 \}$. The distance between $z$ and $z'$ is given by
\begin{equation}
d_\ell(z, z')=arccosh(-\left \langle z, z' \right \rangle_{\mathcal{L} } )    
\label{equ:distance}
\end{equation}
which is also the length of the geodesic that connects $z$ and $z'$. We will refer to the one-hot vector $\mu_0=[1/\sqrt{c},0,0,0...0]\in \mathbb{H}^n \subset \mathbb{R}^{n+1}$ as the origin of the hyperbolic space.


\textbf{Tangent Space of Hyperbolic Space} The tangent space at a point $\mu \in \mathbb{H}^n$ is a Euclidean space composed of vectors. Denoted by $T_{\mu}\mathbb{H}^n$, this tangent space represents the set of vectors in the same ambient space $\mathbb{R}^{n+1}$ where $\mathbb{H}^n$ is embedded. The vectors in $T_{\mu}\mathbb{H}^n$ satisfy an orthogonality condition relative to the Lorentzian product, defined as $T_{\mu}\mathbb{H}^n := \{u: \left \langle \mu, u \right \rangle_\mathcal{L} = 0\}$. This set can be visualized as the tangent space at the point $\mu$ on the forward hyperboloid sheet. Specifically, at the origin $\mu_0$ of $\mathbb{H}^n$, the tangent space $T_{\mu_0}\mathbb{H}^n$ consists of vectors $v \in \mathbb{R}^{n+1}$ . The norm $\left \| v \right \|_\mathcal{L}$, given by the Lorentzian inner product, simplifies to the Euclidean norm $\left \| v \right \|_2$, defined as $\left \| v \right \|_\mathcal{L} := \sqrt{\left \langle v, v \right \rangle_\mathcal{L}} = \left \| v \right \|_2$.

\begin{figure*}
\centering 
\label{framework}
\includegraphics[height=7cm]{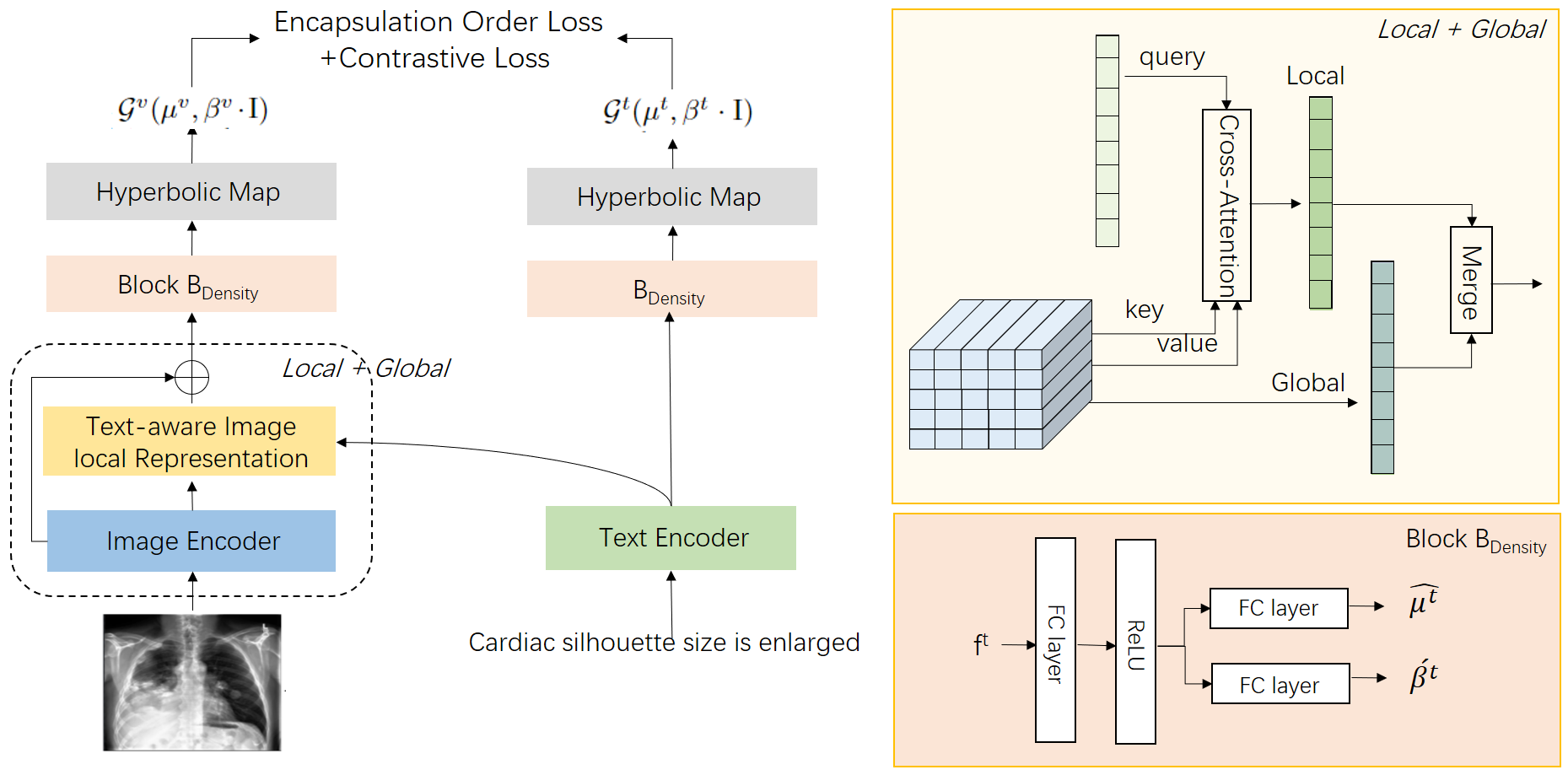}
\caption{
Framework of HYDEN: The contrastive loss function utilizes the negative Lorentzian distance as a metric for similarity. Additionally, an encapsulation loss is employed to enforce the density partial ordering of image and text embeddings within the representation space.
}
\label{fig:framework}
\end{figure*}

\textbf{Exponential Map}

The exponential map provides a method for mapping a vector from a tangent space to its corresponding point on the surface of the hyperbolic space.  For every $u \in T_{\mu}\mathbb{H}^n$,  the exponential map $exp_{\mu}(u):T_{\mu}\mathbb{H}^{n}\to \mathbb{H}^{n}$ allows us to project a vector $u$ in $T_{\mu}\mathbb{H}^{n}$ onto $\mathbb{H}^{n}$ such that the distance from $\mu$ to the destination point of the map coincides with the Lorentzian norm $\left \| u \right \|_\mathcal{L}$ of $u$. In the context of hyperbolic space, the exponential map is given by the equation:

\begin{equation}
z = \exp_{\mu}(u) = \cosh(\left \| u \right \|_\mathcal{L})\mu + \sinh(\left \| u \right \|_\mathcal{L})\frac{u}{\left \| u \right \|_\mathcal{L}}    
\label{equ_projection}
\end{equation}

In this paper, we specifically consider exponential maps where $\mu$ represents the origin of the hyperboloid ($\mathrm {O}=[\sqrt{1/c}, \textbf{0}]$).


\section{Method}

In this section, we present a comprehensive introduction to the HYDEN model. Drawing on the foundation laid by the MERU model \citep{hyperclip} and the widely acclaimed, user-friendly CLIP framework \citep{radford2021clip}, our model adapts and extends these frameworks to address specific challenges in medical image-text representation learning. Figure \ref{fig:framework} depicts the overall architecture of our model. Distinct from CLIP and MERU, our HYDEN model incorporates several innovative features designed to enhance its functionality and applicability in medical contexts: \textbf{(1) Text-aware Local Image Representation:} We refine the image analysis process by integrating text-aware local features that allow for a more nuanced understanding of medical images. This approach helps to align specific image regions with relevant textual descriptions, enhancing the model's ability to handle complex medical datasets. \textbf{(2)Hyperbolic Density Embedding:} Utilizing the properties of hyperbolic space, we introduce hyperbolic density embedding to generate and transfer probability density features from Euclidean to hyperbolic space. This method leverages the natural hierarchical structure of hyperbolic space to more effectively represent the intrinsic complexities of medical data. \textbf{(3)Loss Function for Hyperbolic Density Embedding:} To complement our embedding technique, we have developed a specialized loss function that focuses on the encapsulation relationships within the partial order of image-text semantic distributions. This loss function is tailored to strengthen the correlation between images and texts by capturing their hierarchical semantic relationships.

\subsection{Image-Text Feature Embedding}

In our model, the features $[\hat{f^v}, f^t]$ are derived from respective image and text encoders. For text data, we employ BioClinicalBERT \citep{BioClinicalBERT}, a model that has been pre-trained on the MIMIC III dataset \citep{MIMIC3}, to generate token-level embeddings. Consistent with practices outlined in \citep{PRIOR}, the output of the $[CLS]$ token is used as the medical text feature $f^t$, encapsulating the overall semantic content of the input text.

For image encoding, we utilize the widely-used Vision Transformer (ViT) architecture \citep{vit16}. We assume that $\hat{f^v} = (\hat{f^v_0}, \hat{f^v_1}, \dots, \hat{f^v_n})$ captures the outputs from the image encoder. Recognizing that pathological symptoms often occupy only a portion of a medical image, relying solely on global representations may not adequately capture essential local semantic features. Thus, similar to approaches in \citep{gloria,PRIOR,localrepresentation}, we enhance the global features by integrating text-aware image local representations.

Specifically, we implement a Self-attention module \citep{selfattention}, widely used in cross-modal feature extraction. In this setup, $\hat{f^v}$ acts as both the keys ($K$) and values ($V$), while the text embedding $f^t$ functions as the query ($Q$). This configuration allows us to derive a text-aware image local representation, denoted as $\bar{\hat{f^v}}$. By synthesizing both global and  local representations, we achieve the final image feature $f^v = \hat{f^v} + \bar{\hat{f^v}}$, effectively capturing both the broad and nuanced semantic details present in medical images.

\subsection{Hyperbolic Density Embedding}

Our objective is to transform image-text features into density representations within hyperbolic space. Previous studies such as \citep{Hyperbolicnormal} and \citep{Poincarnormal} proposed methods like the pseudo-hyperbolic Gaussian distribution based on the Lorentz manifold. Due to the computational demands and numerical instabilities of the Poincaré-disk model, we opt for the more stable pseudo-hyperbolic Gaussian distribution for our hyperbolic density embedding. The tangent space \(T_{\mu}\mathbb{H}^n\) of hyperbolic space \(\mathbb{H}^n\) is a Euclidean space, and in \(T_{\mu_0}\mathbb{H}^n\), vectors \(\mathrm{v}\) satisfy \(\mathrm{v} = \{v_0, v_1, ..., v_n\} \in \mathbb{R}^{n+1}\) where \(v_0 = 0\), aligning with the dimensional properties.

To begin, we introduce separate deep nonlinear network blocks, \(B_{density}\), for processing image and text features independently. These blocks do not share parameters, ensuring distinct representations for each modality. As in Figure \ref{fig:framework}, for text features, $\hat{\mu^t}$ and $\acute{\beta^t}$ are the outputs of $B_{density}(f^t)$.

Instead of generating covariance matrices directly, which can introduce numerical instability, we use matrices based on diagonal or spherical assumptions. These are known for their computational efficiency and effectiveness in embedding tasks, particularly in the context of word distribution embedding where spherical covariance matrices have been shown to better model distributional partial order relationships \citep{gaussianembedding}. We thus employ a covariance matrix based on the spherical assumption: $\Sigma^t = \acute{\beta^t} \cdot \mathrm{I} \in \mathbb{R}^{(n+1) \times (n+1)}$.

To ensure that our covariance matrix is positively definite, necessary for the stability of the pseudo-hyperbolic Gaussian distribution, we modify $\beta^t$ using the expression $\beta^t = \text{exp}(\acute{\beta^t})$ referring to solution in VAE\citep{vae}. This adjustment is crucial for maintaining the mathematical integrity of our model when dealing with real-world data. For the embedding vector $\hat{\mu^t} \in \mathbb{R}^{n}$, our aim is to project this vector onto hyperboloid space, which is achieved by mapping it through the exponential function as detailed in Equation \ref{equ_projection}.

The vector $\mu^t_{tan} = [0, \hat{\mu^t}]$ resides in $\mathbb{R}^{n+1}$ and belongs to the tangent space $T_{\mu_0}\mathbb{H}^n$ at the origin of the hyperboloid, $\mathcal{O}$. The norm $||\mu^t_{tan}||_\mathcal{L}$, which equals $||\hat{\mu^t}||_2$, ensures that the mapping preserves the distances inherent to the model's geometric structure. We apply the exponential map to $\mu^t_{tan}$, decomposing the transformation into two parts:
\begin{equation}
    \cosh(\sqrt{c}||\mu^t_{tan}||_{\mathcal{L}})\mathcal{O} = [\sqrt{1/c} \times \cosh(\sqrt{c}||\mu^t_{tan}||_{\mathcal{L}}), \textbf{0}] = [\sqrt{1/c} \times \cosh(\sqrt{c}||\hat{\mu^t}||_2), \textbf{0}]
\end{equation}
\begin{equation}
    \frac{\sinh(\sqrt{c}||\mu^t_{tan}||_{\mathcal{L}})}{\sqrt{c}||\mu^t_{tan}||_{\mathcal{L}}} v_{tan} = [0, \frac{\sinh(\sqrt{c}||\mu^t_{tan}||_{\mathcal{L}})}{\sqrt{c}||\mu^t_{tan}||_{\mathcal{L}}} v_{emb}] = [0, \frac{\sinh(\sqrt{c}||\hat{\mu^t}||_2)}{\sqrt{c}||\hat{\mu^t}||_2} v_{emb}]
\end{equation}

Upon applying the exponential map, we derive the expectation of the hyperbolic density representation:
\begin{equation}
\mu^t = \exp_{\mu_0}(\mu^t_{tan}) = \left(\sqrt{1/c} \times \cosh(\sqrt{c}||\hat{\mu^t}||_2), \frac{\sinh(\sqrt{c}||\hat{\mu^t}||_2)}{\sqrt{c}||\hat{\mu^t}||_2} v_{emb}\right)
\end{equation}

This projection results in the hyperbolic density representation $\mathcal{G}^t(\mu^t, \beta^t \cdot \mathrm{I})$. Following a similar procedure, we also derive $\mathcal{G}^v(\mu^v, \beta^v \cdot \mathrm{I})$ for the image features, thereby ensuring a uniform approach to handling different modalities within our framework.


\subsection{Loss Function Based on Density Embedding}

Traditional point vector embedding often utilizes entailment angle constraints to define relationships between entities \citep{hyperclip}. However, when dealing with probability densities, the notion of partial order can be more complexly captured through the concept of encapsulation. Specifically, a density $f$ is considered more specific than another density $g$ if $f$ is entirely encompassed by $g$, formally expressed as $f \preceq g \Leftrightarrow \{x : f(x) > \eta\} \subseteq \{x : g(x) > \eta\}$, for any $\eta \geq 0$, where $\eta$ indicates the degree of encapsulation necessary for one distribution to entail another.

Imposing such partial order constraints on distributions poses significant challenges. Drawing inspiration from \cite{densityorder}, we employ asymmetric divergence measures between probability densities to address this. We introduce a simple penalty function, $d_\gamma(f, g) = \max(0, D(f \parallel g) - \gamma)$, which serves as a violation penalty rather than as a strict constraint of encapsulation. Here, $D(\parallel)$ represents the divergence measure used to quantify the extent of difference between distributions, and $\gamma$ is a threshold defining the acceptable range of difference.

Among the choices for divergence measures, $\alpha$-divergence provides a more flexible and generalized asymmetric measure \citep{alphadivergence}, allowing for adjustments in the zero-force penalty. This flexibility means that higher $\alpha$ values can enforce stricter encapsulation conditions $f \preceq g$. The general form of $\alpha$-divergence, for $\alpha \neq 0,1$, is given by:
\begin{equation}
D_{\alpha}(f \parallel g) = \frac{1}{\alpha(\alpha-1)} \log \left(\int \frac{f(x)^\alpha}{g(x)^{\alpha-1}} \, dx\right)
\end{equation}

This equation not only quantifies the differences between distributions but also facilitates a deeper understanding of the encapsulation relationships critical for effective density embedding.


We observe that as $\alpha$ approaches 0 or 1, it governs the degree of zero forcing, where minimizing $D_{\alpha}(f \parallel g)$ for high $\alpha$ values results in $f$ becoming more concentrated in regions of $g$ with high density. Conversely, for low $\alpha$ values, $f$ tends to be mass-covering, encompassing regions of $g$ even including those with low density. Notably, there exists a mathematical relationship between KL divergence and $\alpha$-divergence, as indicated by: $\lim_{\alpha \to 1} D_{\alpha}(f \parallel g) = D_{KL}(f \parallel g)$ and $\lim_{\alpha \to 0} D_{\alpha}(f \parallel g) = D_{KL}(g \parallel f)$ \citep{alpha2kl}. Therefore, in our model, we opt for the more flexible and robust $\alpha$-divergence as our metric.

For image-text embedded density $\mathcal{G}^v(\mu^v, \beta^v \cdot \mathrm{I})$ and $\mathcal{G}^t(\mu^t, \beta^t \cdot \mathrm{I})$, the encapsulation loss can be expressed as follows:
\begin{equation}
\begin{split}
d_{\gamma}(\mathcal{G}^v, \mathcal{G}^t) = \max&(0, -\frac{1}{2\alpha(\alpha-1)}\log\Big[\alpha\big(\frac{\beta^v}{\beta^t}\big)^{\alpha(n+1)} \\
&+(1-\alpha)\big(\frac{\beta^t}{\beta^v}\big)^{\alpha(n+1)}\Big]+\frac{(\mu^v-\mu^t)^T(\mu^v-\mu^t)}{\alpha(\beta^t)^{n+1}+(1-\alpha)(\beta^v)^{n+1}} - \gamma)
\end{split}
\end{equation}





Let the batch sample $\mathbb{B} = \{\mathbb{B}^P, \mathbb{B}^N\}$, where $\mathbb{B}^P$ denotes the positive image-text sample set, and $\mathbb{B}^N$ represents the negative set. We define the encapsulation loss function as follows:

\begin{equation}
\mathfrak{L}_{order} = 
\sum_{(\mathcal{G}^t,\mathcal{G}^v)\in \mathbb{B}^P} d_{\gamma}(\mathcal{G}^t, \mathcal{G}^v) +
\sum_{(\mathcal{G}^t,\mathcal{G}^v)\in \mathbb{B}^N} \max\{0, m - d_{\gamma}(\mathcal{G}^t, \mathcal{G}^v)\}
\label{equ:order}
\end{equation}

For the positive samples, a definite partial order relationship exists, enabling the direct application of the density penalty $d_{\gamma}()$. For the negative samples, we enforce the penalty to exceed a margin $m$ due to the absence of an order relationship.

Our goal is to enhance the similarity of semantic distributions between image-text pairs. Therefore, we also employ the classic CLIP contrastive solution \citep{radford2021clip} to compute the geodesic distance between the expectation values of image and text in hyperbolic densities as defined in Equation \ref{equ:distance}, applying Softmax normalization. We define $\mathfrak{L}_{con}$ as the contrastive loss, which is computed as an average of the contrastive losses from both image and text perspectives.

\section{Experiments}

In this section, we aim to rigorously evaluate the performance of our algorithm. We first introduce the baseline model, followed by a description of the medical image-text data and training details used for model pre-training. Then, we discuss the advantages of our proposed model in medical image-text alignment from both quantitative and qualitative perspectives.

A key innovation of our algorithm lies in the use of density representations in hyperbolic space for image-text alignment. To validate the superiority of our approach, we compare it with two methods: CLIP, which aligns image-text pairs in Euclidean space using point embeddings \citep{radford2021clip}, and MERU, which aligns image-text pairs in hyperbolic space using point embeddings \citep{hyperclip}. For the baseline model training, we primarily utilize the open-source code provided by the MERU project\footnote{https://github.com/facebookresearch/meru}. While some variations of CLIP have been successfully applied in the medical image-text alignment domain, our primary focus is on comparing the differences between alignment in Euclidean space and hyperbolic space, as well as between point vector embeddings and distribution embeddings. The reason for retraining the models is that the publicly available CLIP models for the medical domain are mainly intended for downstream fine-tuning tasks and do not include the corresponding text encoder parameters. Since our main objective is to address zero-shot problems, retraining the models on this dataset is necessary.

\begin{table}
    \centering
        \caption{Zero-shot image classification (** 0.01 \& * 0.05 level, delong test)}
    \begin{tabular}{lcccccc}
             \hline
         & \multicolumn{2}{c}{RSNA Pneumonia} & \multicolumn{2}{c}{SIIM-ACR} &\multicolumn{2}{c}{CheXpert 14x100}\\
         \cmidrule(lr){2-3}\cmidrule(lr){4-5}\cmidrule(lr){6-7}
             &     AUC&     F1&   AUC&     F1&  Micro-AUC&  Micro-F1\\

         CLIP&   0.725&  0.489&  0.718& 0.474&  0.583& 0.168\\

         MERU&   0.784&  0.551&  0.754& 0.499&  0.634&  0.2\\

         HYDEN &  $\mathbf{0.845^{**}}$ &  \textbf{0.613}&  $\mathbf{0.778^{*}}$ & \textbf{0.510}&  \textbf{0.642}& \textbf{0.2}\\
                  \hline
    \end{tabular}

    \label{tab:classification}
\end{table}

\begin{table}
    \centering
        \caption{Zero-shot image retrieval}
    \begin{tabular}{lccccccc}
             \hline
         &  Prec@3&  Prec@5&  Prec@10&  NDCG@3&  NDCG@5&  NDCG@10 & Recall@10\% \\
         \hline
         \hline
         & \multicolumn{7}{c}{Text$\longrightarrow$Image}\\

         CLIP&   40.00 & 32.00 & 30.00 & 0.588 & 0.588 & 0.576 & 4.63\% \\

         MERU&   37.78 & 34.67 & 29.33 & 0.534 & 0.616 & 0.635 & 4.97\% \\

         HYDEN &  \textbf{46.67}&\textbf{41.33}&  \textbf{40.67}&  \textbf{0.676}&  \textbf{0.723}&  \textbf{0.736}& \textbf{4.2\% }\\
         \hline          \hline
                  & \multicolumn{7}{c}{Image$\longrightarrow$Image}\\
        
         CLIP&   26.67&22.67&24.00&0.603&0.598&0.600&6.79\% \\

         MERU&   33.33&32.00&26.00&0.484&0.478&0.463&6.99\% \\

         HYDEN &  \textbf{37.78}&\textbf{38.67}&  \textbf{34.67}&  \textbf{0.659}&  \textbf{0.765}&  \textbf{0.766}& \textbf{4.30\% }\\

                  \hline
    \end{tabular}
    \label{tab:retrive}
\end{table}

\subsection{Training Details}

\textbf{Datasets:} We train our alignment model using the MIMIC-CXR v2 dataset \citep{mimic}, comprising over 227,000 studies of paired image-report data sourced from 65,379 patients undergoing various scans. Each study may contain one or two images, representing different scan views, resulting in a total of 377,110 images. During training, we perform random cropping, flipping, rotation, and other data augmentation techniques on the images, while also resizing them to a [224,224] dimension. Additionally, for the text data, we augment the reports by randomly adding medical entity prefixes to enhance semantic information, such as '{event\_list}: {report}'.

\textbf{Settings:} We employ ViT-B \citep{vit16} with a patch size of 16 as the image encoder, as it has demonstrated competitive performance in hyperbolic space \citep{hyperclip}. Our initialization strategy for image/text encoders follows a similar style to MERU, with the exception of utilizing ClinicalBERT \citep{clinicalbert} as the pre-trained text encoder, which has been pre-trained on large-scale medical text data. For HYDEN, we initialize the learnable curvature parameter $c$ to 1.0 and clamp it within the range of [0.1, 10.0] to prevent training instability. All experiments were conducted using two NVIDIA A40 GPU and the PyTorch framework


\textbf{Optimization:} We adopt the AdamW optimizer with a weight decay of 0.2 and $(\beta_1, \beta_2) = (0.9, 0.98)$. Weight decay is disabled for all gains, biases, and learnable scalars. Models are trained for 13,000 iterations with a batch size of 256. The maximum learning rate is set to $1\times 10^{-5}$, linearly increased for the first 500 iterations, followed by cosine decay to zero. We leverage mixed precision to expedite training, except when computing exponential maps and losses, where FP32 precision is used for numerical stability.

\subsection{Quantitative Analysis}
We evaluate all baselines and HYDEN on three categories of zero-shot downstream tasks, classification, text-image retrieval and image-image retrieval. We use three public datasets for the evaluation, where both \textbf{RSNA Pneumonia} \citep{RSNA_Pneumonia} and \textbf{SIIM-ACR Pneumothorax} \citep{SIIM-ACR} are used for binary classification, \textbf{ChestXray14}\citep{chestxray14} is used for multi-label classification, text-image retrieval and image-image retrieval. For the two binary classification tasks, we report the Area Under the Curve (AUC) and F1 score; for the multi-label task, we provide the Micro-AUC and Micro-F1. For the retrieval task, Top-k Precision (abbreviated as Prec@k) and Tok-k Normalized Discounted Cumulative Gain (abbreviated as NDCG@k) are used to evaluate the retrieval performance. Refer Appendix B for details about our evaluation tasks and datasets.

\textbf{Zero-shot Image Classification} Table \ref{tab:classification} presents the performance of the baselines and HYDEN across three classification datasets. The results indicate that HYDEN consistently demonstrates robust transfer classification performance, both in binary classification tasks and multi-label classification task. Compared to CLIP, both MERU and HyperMed achieved improved accuracy. This suggests that using hyperbolic space for text-image representations, especially for medical data characterized by a visual semantic hierarchy, is more effective. Relative to MERU, HYDEN achieved the highest accuracy across almost all of metrics, highlighting the advantages of density embedding-based representation methods over point vector embedding, particularly in addressing the challenges of semantic uncertainty.

\textbf{Zero-shot Retrieval} Table \ref{tab:retrive} displays the performance of two baseline models and HYDEN in "image-to-text" and "image-to-image" retrieval tasks. The results demonstrate that representation learning in hyperbolic space mostly outperforms that in Euclidean space; among the methods, HYDEN exhibits the best retrieval performance. Furthermore, we observed a significant enhancement in the ranking quality of HYDEN's retrieval results compared to the two baseline methods. We hypothesize that this improvement is linked to the method of density embedding. Similar to findings in the recommendation systems domain \citep{gaussiancf}, unlike point vector embeddings, density embeddings enable better handling of uncertainties, information sparsity, ambiguity, and even contradictions, which are common challenges in medical image-text data.

\begin{table}
\centering
\caption{Ablation Study of HYDEN: This table presents the results of ablating three key design choices within the HYDEN framework to evaluate their individual contributions.}
\begin{tabular}{lcccccccc}
\hline
& \multicolumn{2}{c}{Text2Image@10} & \multicolumn{2}{c}{Image2Image@10} &  \multicolumn{2}{c}{RSNA} & \multicolumn{2}{c}{SIIM} \\
& Prec & NDCG & Prec & NDCG & AUC & F1 & AUC & F1\\
\hline
HYDEN              & 40.67 & 0.736  & 34.67  & 0.766 & 0.845 & 0.613 & 0.778 & 0.510 \\
1. w KL Divergence & 34    & 0.609  & 31.33 &  0.642 & 0.842 & 0.609 & 0.759 & 0.466 \\ 
2. w/o encapsulation loss    & 19.33  & 0.384  & 16.67 & 0.416 & 0.669 & 0.503 & 0.663 & 0.440\\
3. w/o local representation  & 20.67  & 0.532  & 34    & 0.643 & 0.772 & 0.566 & 0.768 & 0.434\\
\hline
\end{tabular}
\label{tab:ablation}
\end{table}

\begin{table}
\centering
\caption{Impact of Different $\alpha$ Settings on $\mathfrak{L}_{order}$ Variant}
\begin{tabular}{lcccccccc}
\hline
& \multicolumn{2}{c}{Text2Image@10} & \multicolumn{2}{c}{Image2Image@10} &  \multicolumn{2}{c}{RSNA} & \multicolumn{2}{c}{SIIM} \\
& Prec & NDCG & Prec & NDCG & AUC & F1 & AUC & F1\\
\hline
$~~~~\alpha=0.6$ & 31.33 & 0.566  & 30.0  & 0.653 & \textbf{0.853} & 0.609 & 0.746 & 0.457 \\
$~~~~\alpha=0.7$ & \textbf{40.67} & \textbf{0.736}  & \textbf{34.67} & \textbf{0.766} & 0.845 & 0.613 & 0.778 & \textbf{0.510} \\
$~~~~\alpha=0.8$ & 31.33 & 0.703  & 29.33 & 0.619 & 0.845 & \textbf{0.614} & \textbf{0.788} & 0.462 \\
$~~~~\alpha=0.9$ & 29.33 & 0.592  & 32.0  & 0.679 & 0.800 & 0.576 & 0.784 & 0.474 \\

\hline
\end{tabular}
\label{tab:alpha}
\end{table}
\textbf{Ablation Studies}
In this section, we examine the impact of different design choices using HYDEN. Specifically, we trained three ablation models with default hyperparameters, and the results are presented in Table \ref{tab:ablation}. From Table \ref{tab:ablation}, we observe that: (1) Using $\alpha$-divergence in the loss function instead of KL divergence better aligns with the encapsulation's partial order properties of text-image distribution embeddings. The experimental results also indicate that replacing $\alpha$-divergence with KL divergence leads to performance degradation across all tasks. (2) Omitting the encapsulation loss, i.e., not using $\mathfrak{L}_{order}$ as defined in Equation \ref{equ:order} and relying primarily on $\mathfrak{L}_{con}$, results in performance degradation across all tasks. This is because not using encapsulation loss implies that the prior partial order of text and image cannot be utilized in hyperbolic space, thus losing the benefits introduced by hyperbolic geometry. (3) The model experiences a performance drop across all tasks when not using text-aware image representation. This is primarily due to the nature of medical image-text features. As discussed in the Introduction, most regions in medical images may differ in texture and morphology but not in clinical significance, while actual pathological changes are localized. The results also show that enhancing text-aware local features is meaningful for medical image-text alignment.

\textbf{Parameters Impacts}
In this section, we focus on adjusting the crucial predefined parameter $\alpha$ within the HyperMed model to observe its impact on the results. The experimental outcomes are displayed in Figure \ref{tab:alpha}. The table reveals that different $\alpha$ values directly influence the evaluation outcomes. The performance does not continually increase with larger $\alpha$ values; instead, it exhibits a quasi-convex distribution, initially increasing and then decreasing, which may be related to the distribution of the training data. In the experiments mentioned above, we employ a hyperparameter setting of $\alpha=0.7$.

\subsection{Qualitative Analysis}
In this section, we explore the trained models to deduce the characteristics of the model in capturing the visual semantic hierarchy structure. The concept of 'Embedding distances from [ROOT]' was introduced by \cite{hyperclip} to depict the generality differences between text and image embeddings in hyperbolic space. This concept highlights that in a representation space that effectively captures the visual semantic hierarchy, text embeddings are typically more general than image embeddings and, therefore, should be closer to the root node [ROOT].

\begin{figure}[htbp]	
	\subfigure[CLIP] 
	{
		\begin{minipage}{4.4cm}
			\centering          
			\includegraphics[width=4.4cm]{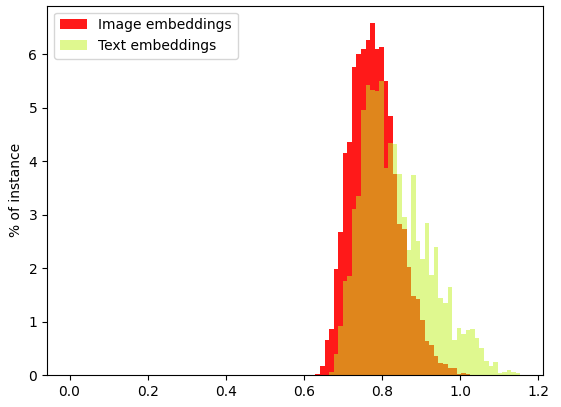}   
		\end{minipage}
	}
	\subfigure[MERU] 
	{
		\begin{minipage}{4.4cm}
			\centering      
			\includegraphics[width=4.4cm]{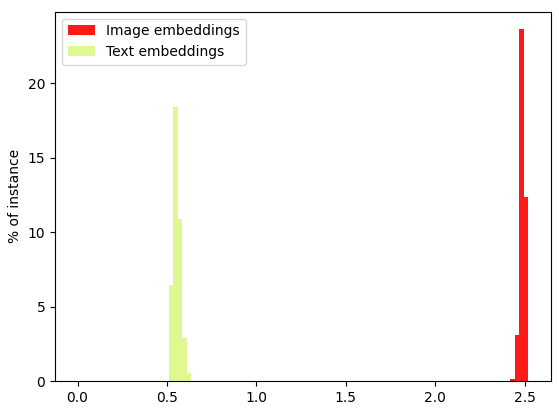}   
		\end{minipage}
	}
        \subfigure[HYDEN] 
	{
		\begin{minipage}{4.4cm}
			\centering      
			\includegraphics[width=4.4cm]{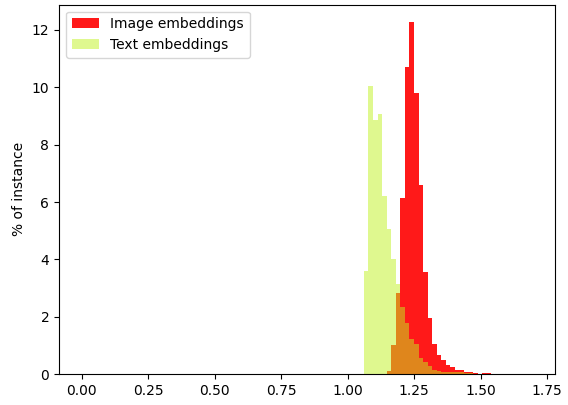}   
		\end{minipage}
	}
\caption{
Distribution of embedding distances from [ROOT]: We embed 3858 testing images and text from the MIMIC-CXR v2 dataset using pre-trained CLIP, MERU, and HYDEN models.
}
    \label{fig:dist_distribution}  
\end{figure}

Here, we visualize the differences in distance distributions between text and image embeddings. Given that our approach utilizes distribution embeddings, we specifically visualize the expectations of the distance distributions of text and image density embeddings. Figure \ref{fig:dist_distribution} demonstrates that the distribution differences generated by our model lie between those produced by MERU and CLIP, with some overlapping distribution areas. This suggests that our model is capable of capturing the visual semantic hierarchy. Compared to the diversity of natural text, medical image-text data is relatively uniform, and the introduction of distributions diminishes the effect of prior entailments, explaining why our model achieves significantly higher accuracy and ranking quality in both text-image and image-image retrieval tasks.



\section{Conclusion}
In this paper, we propose a novel approach, HYDEN, to text-image representation learning based on hyperbolic density embeddings. It is a representation learning method tailored for specific medical domain data. Experimental results demonstrate the interpretability of our method and its superior performance compared to baseline methods across various zero-shot tasks and different datasets.

\textbf{Limitations.} Our work is not without limitations. While our method performs well in zero-shot retrieval and image classification tasks, it cannot be directly applied as a pre-trained model to downstream fine-tuning tasks. This is because downstream fine-tuning tasks mainly involve classification, segmentation, recognition, etc., based on Euclidean space. Applying our model to other tasks involving few-shot learning or full-model fine-tuning is also beyond the scope of this paper.


\bibliographystyle{unsrtnat}
\bibliography{main}
\newpage
\appendix

\section{Material}
\textbf{R$\acute{e}$nyi $\alpha$-Divergence} is a general family of divergences that introduce varying degrees of zero-forcing penalty. The general form of the $\alpha$-divergence for $\alpha \ne 0,1$ is described as below,
\begin{equation}
D_{\alpha}(f||g) = \frac{1}{\alpha(\alpha-1)}\log\left(\int \frac{f(x)^{\alpha}}{g(x)^{\alpha-1}} , dx\right)\end{equation}
It is notable that as $\alpha$ approaches 0 or 1, the $\alpha$-divergence converges to the KL divergence and the reverse KL divergence, respectively. For two multivariate Gaussians $f$ and $g$, the Rényi $\alpha$-Divergence can be expressed as:
\begin{equation}
D_{\alpha}(f||g)=-\frac{1}{2\alpha(\alpha-1)}log\frac{det(\alpha\Sigma_g+(1-\alpha)\Sigma_f)}{(det(\Sigma_f)^{1-\alpha}\cdot det(\Sigma_g)^{\alpha})}+(\mu_f-\mu_g)^T (\alpha\Sigma_g+(1-\alpha)\Sigma_f)^{-1}(\mu_f-\mu_g)  
\end{equation}
Here, the parameter $\alpha$ modulates the extent of zero forcing: minimizing $D_{\alpha}(f||g)$ for high $\alpha$ values results in $f$ being concentrated towards the high-density regions of $g$. Conversely, for low $\alpha$, $f$ tends to have broader support, covering regions of $g$ including those with low density.

\section{Evaluation Tasks \& Data}

\textbf{Zero-shot Image Classification:} We evaluate the pre-trained model on three representative medical image classification tasks:

\begin{enumerate}
\itemsep0em
    \item \textbf{RSNA Pneumonia Dataset}\citep{RSNA_Pneumonia}: Comprising over 260,000 frontal chest radiographs collected by the Radiological Society of North America (RSNA). These images can be classified into a binary classification task: pneumonia vs. normal. For evaluation purposes, we randomly sample 4003 images for evaluation. 
    
    \item \textbf{SIIM-ACR Pneumothorax Dataset}\citep{SIIM-ACR}: Contains more than 12,000 frontal chest radiographs collected by the Society for Imaging Informatics in Medicine and the American College of Radiology (SIIM-ACR). Similar to the RSNA Pneumonia dataset, it is used for a binary classification task to determine the presence or absence of pneumothorax. We use all 10,675 images for evaluation.
    
    \item \textbf{ChestXray14 Dataset}\citep{chestxray14}: NIH ChestXray14 has 112,120 chest X-ray images with 14 disease labels from 30,805 unique patients. The official test set released by the NIH, comprising 22,433 images, are distinctively annotated by board certified radiologists. For multi-label evaluation, we only test on the official test set.
\end{enumerate}

\textbf{Zero-shot Text-Image Retrieval:}
For pre-training methods akin to CLIP, text-image retrieval tests are standard practice.  Following the practices of CLIP \citep{radford2021clip} and MERU \citep{hyperclip}, we also introduce downstream tasks for text-image retrieval. In medical imaging reports, the same diagnosis often has varied textual descriptions, making retrieval from image to text impractical. Thus, we do not use images to query text; instead, we use text to retrieve specific categories of images as described in \citep{convirt}. For this purpose, we first construct a text-image retrieval evaluation dataset. As described in the multi-label classification task, ChestXray14 \cite{chexpert14} encompasses 14 different disease classes and one 'normal' class, totaling 15 categories. Based on these class labels, we randomly extract 100 images for each class (exclusive), forming the ChestXray14x100 dataset, which consists of 1,500 images. We then write representative text prompts for each of the 15 categories. During testing, for each query, we encode its text using the learned text encoder, then retrieve from the candidate images in a similar manner. This evaluation assesses not only the quality of the learned image representations but also the consistency between text and image representations. 


\textbf{Zero-shot Image-Image Retrieval:} This evaluation task is similar to traditional content-based image retrieval tasks and is also a common downstream task in medical imaging. It involves using a query image to search for images of specific categories. To evaluate, a set of query images, image category label prompts, and a candidate image set are provided to the pre-trained representation model. We encode each query and candidate image using the encoder of the pre-trained model, then rank all candidate images in descending order of their reciprocal geodesic distance from the query's expected distribution. 

\textbf{Prompts Design:} \\
To create the textual queries for each category on each evaluation task, we consulted a board-certified radiologist to draft at least five distinct sentences describing each abnormality as it would appear in radiology reports. Drawing inspiration from ConVIRT\citep{convirt}, we established the following criteria: 1) The sentences must clearly describe the specific category without ambiguity and should not reference other categories.
2) The sentences must be varied and distinct from one another.
3) The sentences should avoid mentioning highly specific anatomical locations or rare clinical findings.

Finally, we aggregated the results and selected the best textual queries for each abnormality category. For reference, examples of the textual queries are presented in Table \ref{tab:prompts}.

\begin{table}
\centering
\caption{Example prompts for each categories in the evaluation tasks.}
\begin{tabular}{ll}
\hline
\textbf{Disease Category} & \textbf{Prompts for RSNA Pneumonia Tasks} \\
\hline
 Normal & The chest image can not find any symptoms. \\
 Pneumonia& The chest image shows the pneumonia. \\
\hline
\hline
\textbf{Disease Category} & \textbf{Prompts for SIIM Pneumothorax Tasks} \\
\hline
 Normal & The chest image can not find any symptoms. \\
Pneumothorax & The chest image shows the pneumothorax. \\
\hline
\hline
\textbf{Disease Category} & \textbf{Prompts for Text to Image Retrieval Tasks} \\
\hline
Atelectasis&A subtle opacity in the lung base could be attributed to a patch of atelectasis.\\
Cardiomegaly&The cardiac silhouette is prominently enlarged, pointing to possible cardiomegaly.\\
Effusion&Fluid levels observed within the pleural cavity.\\
Infiltration&Hazy densities throughout the lung parenchyma, indicative of infiltration.\\
Mass&A mass lesion is noted, warranting further evaluation.\\
Nodule&A solitary small pulmonary density suggestive of a nodule.\\
Pneumonia&Airspace disease with lobar distribution points to possible pneumonia.\\
Pneumothorax&The chest film shows pneumothorax with lung collapse.\\
Consolidation&Areas of dense opacity suggest alveolar consolidation.\\
Edema&Pulmonary edema is suggested by perihilar haziness.\\
Emphysema&Lung parenchyma shows large areas of low attenuation, suggesting emphysema.\\
Fibrosis&Linear and nodular opacities indicative of lung fibrosis.\\
Pleural Thickening&The pleural surfaces show signs of fibrotic changes, suggesting pleural thickening.\\
Hernia&There is evidence of a diaphragmatic hernia.\\
No Finding&The chest radiograph shows no abnormality.\\
\hline
\hline
\textbf{Disease Category} & \textbf{Prompts for Image to Image Retrieval Tasks} \\
\hline
Atelectasis&Linear areas of opacity are consistent with areas of atelectasis.\\
Cardiomegaly&Cardiomegaly is indicated by an increased cardiothoracic ratio.\\
Effusion&There is fluid accumulating in the pleural space indicative of pleural effusion.\\
Infiltration&The presence of diffuse lung markings suggests pulmonary infiltration.\\
Mass&An abnormal density is identified, consistent with a mass lesion.\\
Nodule&A well-defined rounded opacity suggests the presence of a pulmonary nodule.\\
Pneumonia&An area of consolidation with air bronchograms indicates pneumonia.\\
Pneumothorax&The presence of free air in the pleural space suggests a pneumothorax.\\
Consolidation&Consolidation is suspected due to a region of lung opacification.\\
Edema&Pulmonary edema is suggested by perihilar haziness.\\
Emphysema&Hyperinflation and flattened diaphragms suggest emphysema.\\
Fibrosis&Fibrosis is indicated by reticular opacities in the lung fields.\\
Pleural Thickening&The pleura appears thickened, indicating pleural thickening.\\
Hernia&An organ protrusion through the diaphragm suggests a hernia.\\
No Finding&The imaging shows no significant abnormalities.\\

\hline

\end{tabular}
\label{tab:prompts}
\end{table}

\end{document}